# Application and Evaluation of Large Language Models for Forecasting the Impact of Traffic Incidents

George Jagadeesh*, Srikrishna Iyer, Michal Polanowski and Kai Xin Thia

*Abstract*— This study examines the feasibility of applying large language models (LLMs) for forecasting the impact of traffic incidents on the traffic flow. The use of LLMs for this task has several advantages over existing machine learning-based solutions such as not requiring a large training dataset and the ability to utilize free-text incident logs. We propose a fully LLM-based solution that predicts the incident impact using a combination of traffic features and LLM-extracted incident features. A key ingredient of this solution is an effective method of selecting examples for the LLM's in-context learning. We evaluate the performance of three advanced LLMs and two state-of-the-art machine learning models on a real traffic incident dataset. The results show that the best-performing LLM matches the accuracy of the most accurate machine learning model, despite the former not having been trained on this prediction task. The findings indicate that LLMs are a practically viable option for traffic incident impact prediction.

## I. INTRODUCTION

Traffic incidents such as accidents are a leading cause of non-recurring congestion, which imposes substantial costs on travelers in terms of wasted time, productivity and fuel. While many traffic incidents do not significantly affect the traffic flow, some cause substantial disruptions. An accurate forecast of a just-occurred traffic incident's future impact on traffic is valuable for travelers (to avoid congested routes) and traffic managers (to respond appropriately to the incident). However, due to the stochastic nature of traffic incidents, it is generally hard to predict how an ongoing incident would affect the traffic at a future time point.

Machine learning models have been previously proposed for predicting different aspects of traffic incidents' impact such as the incident duration [1], the impacted spatial span [2], and the incident-induced delay [3]. However, the use of machine learning models for traffic incident impact prediction suffers from two disadvantages. First, in order to train such models, a labelled dataset containing data of at least a few hundred incidents from the target region is generally required. Second, valuable information about an ongoing traffic incident is generally available in unstructured textual format (e.g., messages from responders), which cannot be readily utilized by the machine learning models.

In the last few years, pre-trained large language models (LLMs) are increasingly being used to solve prediction problems in intelligent transportation systems (ITS) [4]. In the context of predicting the impact of traffic incidents, LLMs offer several advantages. They are capable of in-context learning (ICL), which is the ability to perform new tasks based on a few examples provided in the prompt, without the need to retrain the model with a large labelled dataset. LLMs, of course, have an impressive ability to process unstructured texts and extract relevant information. Furthermore, unlike conventional prediction models, LLMs can use their pre-learned knowledge about the given problem (e.g., how different aspects of traffic incidents affect the traffic flow).

Our main goal in this paper is to investigate the effectiveness of state-of-the-art LLMs for forecasting the impact of traffic incidents. Our work towards this goal includes LLM-based extraction of incident features from unstructured text logs, applying multiple LLMs for traffic incident impact prediction through optimal use of ICL, and comparative evaluation of the LLM-based prediction models and conventional machine learning-based methods on a real traffic incident dataset. The main contributions of the paper are as follows:

- We present a fully LLM-based solution for predicting the impact of traffic incidents. It uses a small-scale LLM for extracting numeric features from unstructured text logs. In order to optimally leverage the ICL ability of LLMs, we have devised a simple yet effective method of selecting examples to be included in the prompt from a set of labelled incidents.

- We evaluate the prediction accuracies of three advanced LLMs (Claude 3.7 Sonnet, Gemini 2.0 Flash and GPT 4.1) on a dataset with 556 labelled traffic incidents and compare them with two popular machine learning models. Despite relying on a relatively small number of examples, the LLM-based solutions achieve accuracy comparable to the machine learning models trained with a large labelled dataset. Furthermore, the experiments confirm the effectiveness of the proposed method of selecting examples for ICL.

## II. RELATED WORK

Prediction problems related to traffic incidents are challenging to solve due to the high degree of variability involved. However, such problems have been the subject of active research in ITS. For the last 10 years or so, machine learning models have been the conventional approach for predicting the impact of traffic incidents, especially their durations [1]. A study by Mihaita et al. [5] found XGBoost and Random Forest to be the two best performing models for incident duration prediction. More recently, some researchers have applied deep learning techniques such as graph neural networks and long short-term memory for predicting the

*Corresponding author

G. R. Jagadeesh, S. Iyer, M. Polanowski and K. X. Thia are with the Artificial Intelligence – Data Analytics Strategic Technology Centre, ST Engineering Ltd., 600 West Camp Road, Singapore 797654. (e-mail: {george.jagadeesh, srikrishna.rameshiyer, michal.polanowski, kaixin.thia} @stengg.com).

impact of traffic incidents in terms of the impacted spatial span (i.e., the queue length) [2][6].

Pereira et al. [7] showed that when features extracted from free-text incident reports are combined with other features, they consistently improve the accuracy of traffic incident duration prediction. Before LLMs became prevalent, natural language processing (NLP) techniques have been used to extract features from unstructured incident reports [8][9]. Some researchers [10][11] have used early LLMs such as bidirectional encoder representations from transformers (BERT) to encode unstructured incident reports into meaningful representations (embeddings), which are then fed to machine learning models to predict the duration or severity of traffic incidents. To our knowledge, the direct use of LLMs for traffic incident impact prediction has so far not been explored in the literature despite the potential benefits of such an approach.

## III. PRELIMINARIES AND PROBLEM DEFINITION

### A. Traffic Incident Impact

With the widespread deployment of traffic sensors, it has become feasible to empirically determine the impact of traffic incidents based on traffic data. A number of such empirical methods have been proposed in the literature (e.g., [12] [13]). Most of them determine the incident impact at a post-incident time point based on the decrease in speed with respect to the normal speed at the time. Our method of determining the incident impact in this paper is also based on the decrease in speed but we differ in some aspects.

We calculate the impact of a traffic incident at a post-incident time step based on the decrease in speed with respect to the speed prevailing before the incident along a 2-mile road stretch upstream of the incident location. We consider the pre-incident speed to account for any non-recurring congestion already existing when the incident occurred.

In our work, time is divided into 5-minute time steps. Let the *relative speed* of a traffic sensor $s$ any time step $t$, denoted as $v_r(s,t)$, be the difference between the measured speed and the historical average speed for that time of the day. We calculate the *pre-incident relative speed* for an incident $i$, denoted as $\rho(i)$, as the average of the relative speeds of all the sensors along the 2-mile upstream stretch in the 3 time steps before the incident. For any upstream sensor $s$ at any post-incident time step $t$, we calculate the *speed decrease ratio* as

$$\delta_v(s,t) = max\left(\frac{\rho(i) - v_r(s,t)}{\hat{v}(s,t)}, 0\right) \quad (1)$$

where $\hat{v}(s,t)$ is the corresponding historical average speed. The overall speed decrease ratio for incident $i$ at post-incident time step $t$, denoted as $\Delta_v(i,t)$, is the average of the speed decrease ratios of all the sensors in the upstream stretch.

Based on the overall speed decrease ratio $\Delta_v(i,t)$, we determine the impact of the traffic incident $i$ at post-incident time step $t$ as one of three classes, namely, mild, moderate and severe. They are defined as follows:

Mild:        $\Delta_v(i,t) \leq 0.2$
Moderate:   $0.2 > \Delta_v(i,t) \leq 0.5$
Severe:       $\Delta_v(i,t) > 0.5$

### B. Problem Definition

We model the task of predicting the impact of a traffic incident as a classification problem where the target variable is one of the three classes defined above. We do not attempt to predict the incident impact as a precise numerical value as doing so has been found to yield poor prediction accuracy [8]. Furthermore, a qualitative assessment of the incident impact as mild, moderate or severe is meaningful and adequate for decision making in most scenarios. The prediction is performed based on the features of the incident and features derived from traffic speed measurements observed till the prediction time.

We define the problem as follows: Given the attributes of a traffic incident $i$ that occurred at time step $j$ and the traffic speed measurements upstream of the incident location up to time step $j$, predict the impact of the incident on the upstream traffic at a subsequent time step $k$, where $k > j$.

## IV. LLM-BASED TRAFFIC INCIDENT IMPACT PREDICTION

Fig. 1 shows the proposed LLM-based solution for traffic incident impact prediction. The main aspects of the solution are discussed below.

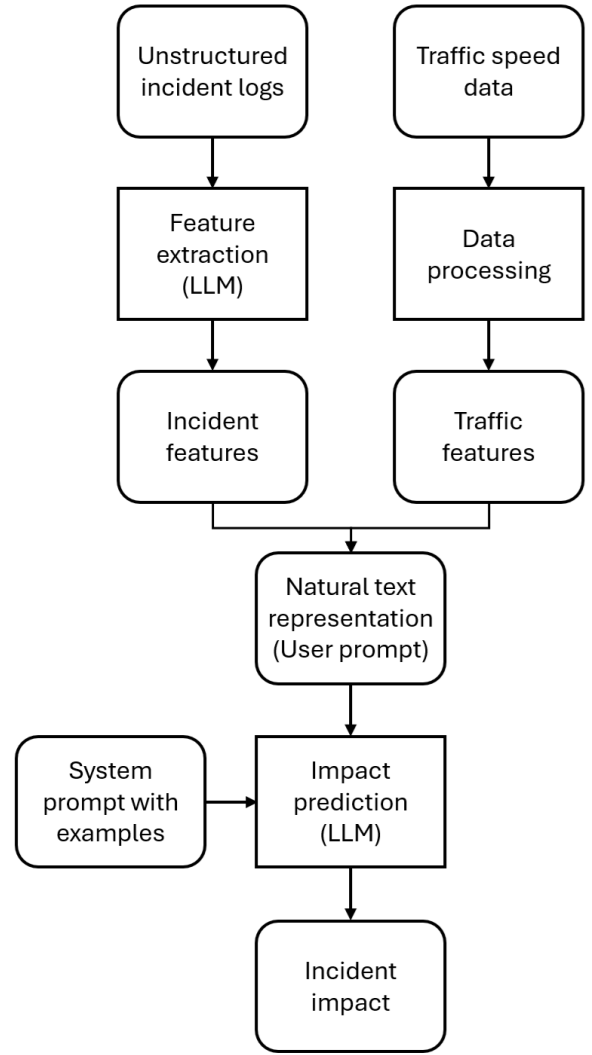

Figure 1. The LLM-based solution for traffic incident impact prediction

## A. Feature Extraction

In the first stage of the workflow, incident features and traffic features are extracted from unstructured incident logs and traffic speed data, respectively. A small-scale LLM is used to process the free-flow incident logs and extract a number of features such as the number of vehicles involved, the number of lanes blocked, if a large truck is involved, if an ambulance / tow truck has been called, if there is any mention of adverse weather etc. However, as described in Section V, it was later found that only the following incident features are actually useful for impact prediction:

- The time of the incident
- The number of vehicles involved in the incident
- The number of lanes blocked due to the incident

The traffic speed data up to the prediction time is separately processed to calculate the following traffic features as described in the previous section:

- The pre-incident relative speed
- The overall speed decrease ratio at the prediction time

The pre-incident relative speed is an indicator of how the traffic conditions were before the incident, compared to the normal traffic conditions at that time of the day. The overall speed decrease ratio at the prediction time represents the initial decrease in the upstream traffic speeds from the time of the incident up to the prediction time. It needs to be clarified that the impact prediction is not done at the exact time an incident is reported but at the end of the 5-minute time step. For example, if an incident was first reported at 8:08 AM, the impact prediction would be done at 8:10 AM.

## B. Impact Prediction

For a given incident, the above-listed incident features and traffic features are combined into a natural text representation and supplied to an LLM as a user prompt. An example of such a prompt corresponding to a real incident in the dataset used in this study is given below:

*A traffic collision incident occurred at 4:49 PM. Three vehicles are involved. One lane is blocked currently. The pre-incident traffic speed was 12 mph below the historical mean speed for the time of the day. The initial decrease in speed in the first few minutes after the incident is 8.75%. Predict what would be the impact after 15 minutes.*

Besides the user prompt, a system prompt is also provided to the LLM. The system prompt instructs the LLM how it should process the user prompt and respond. It contains additional information about the incident impact prediction task including a definition of the impact classes and some common knowledge about how different aspects of a traffic incident affect the upstream traffic speeds. (e.g., "A higher number of blocked lanes may probably cause a bigger decrease in speed.") The LLM is also instructed through the system prompt to provide its prediction as a single word: mild, moderate or severe. More importantly, in order to enable the LLM's ICL capability, the system prompt contains several examples that are instances of correct predictions.

## C. Selection of Examples for In-Context Learning

An LLM's ability to perform novel tasks through ICL depends very much on the choice of the examples provided to it to show how the task should be done correctly [14]. The common practice of randomly selecting a few examples from a labeled set and including them in a prompt results in poor accuracy for classification tasks [14]. Against this background, we have designed a simple yet effective method of automatically selecting examples from a set of labelled traffic incidents.

Fig. 2 provides an overview of the proposed method. It involves selecting a small number $m$ of examples from the labelled set of incidents according to a judicious sampling strategy (described below) to construct a prompt and evaluating the prompt on a validation set using an LLM. The validation set is a small subset of the labelled set. 30 different prompts each with $m$ examples are evaluated against the validation set. The top $k$ prompts with the best prediction accuracies are selected and combined to form the final prompt with $k \times m$ examples. The values of $k$ and $m$ are to be determined empirically.

*The sampling strategy*: The first step in the sampling strategy is to identify and exclude outlier incidents belonging to each class. An incident is considered an outlier if it is closer to the centroid of another class in the normalized feature space than to the centroid of its own class. In the second step, for each class, a subset (50%) of non-outlier incidents that are closest to its neighbor class are identified. (We assume that the mild and severe classes, respectively, have only one neighbor class, the moderate class. The latter has two neighbor classes.) We refer to this subset of non-outlier incidents as the near-boundary incidents, which probably would help the LLM to better classify edge cases. In the final step, for each of the 3 classes, $m/3$ incidents are randomly selected from among the near-boundary incidents. Thus, a total of $m$ incidents are selected as examples to be included in the prompt.

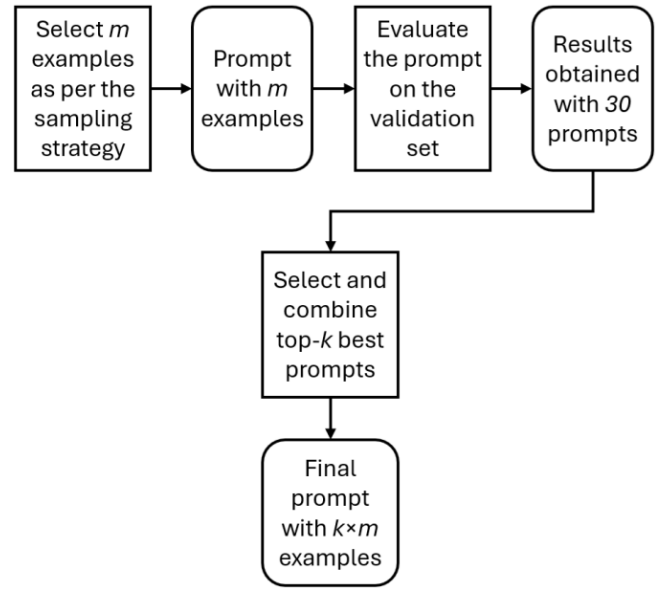

Figure 2. Overview of the proposed example selection method

## V. EXPERIMENTS

### A. Dataset and Data Preparation

We have prepared a dataset consisting of 2777 traffic incidents corresponding to the PEMS-BAY traffic dataset [15], which contains traffic speed data from 325 freeway sensors in the San Francisco Bay Area for the first 6 months of 2017. From the traffic incident data available on the Caltrans Performance Measurement System (PeMS) website [16], we have selected the incidents that are classified as accidents. Out of these, we have excluded the incidents that occurred within the first 2 miles of each freeway as it is not possible to determine the ground-truth impact on the upstream traffic for such incidents. Each incident is associated with a record of unstructured time-stamped messages logged by the California Highway Patrol (CHP) during the course of the incident.

We have marked the first (i.e., earliest) 80% of the incidents (2221 incidents) as a training set, which is used mainly for training the machine learning models that are used as comparison references. The remaining 556 incidents are marked as a test set for evaluating all the impact prediction models. Using the traffic speed data from the PEMS-BAY dataset, we have determined the ground-truth impact class of each incident for two prediction horizons (15 minutes and 30 minutes) as explained in Section III. As shown in Table I, most incidents have a mild impact on the traffic with only a small proportion of them having a moderate or severe impact.

For each incident, we have determined the two traffic features (the pre-incident relative speed and the overall speed reduction ratio at prediction time) using the traffic speed data. To extract the incident features, we have first preprocessed the incident logs to replace known abbreviations and codes based on a glossary made available by CHP. Subsequently, we have used a small-scale LLM (GPT-4o mini) to analyze the incident logs generated up to the prediction time and extract a number of incident features. Our examination of a sampling of the incident logs and the corresponding extracted features indicate that the LLM is able to extract the features with human-level correctness.

### B. Experimental Setup

For each incident in the test set, impact prediction is done at the end of the 5-minute time step during which the incident is first reported. Predictions are made for two prediction horizons, 15 minutes and 30 minutes.

We evaluate the following state-of-the-art LLMs on the task of traffic incident impact prediction: Claude 3.7 Sonnet, Gemini 2.0 Flash and GPT 4.1. The LLMs are accessed through API calls with the temperature parameter set to 0 to make the output more deterministic and consistent. For the method of selecting examples for ICL discussed in Section IV, we use a small subset of the training set as the validation set. The validation set consists of 60 incidents with 20 random samples for each class. The examples to be included in the prompts are selected from the rest of the training set as per the sampling strategy. The values of $m$ and $k$ in the example selection method are empirically set to 12 and 2, respectively, so that the final system prompt includes 24 (i.e., $k \times m$) examples (8 examples for each class).

TABLE I. THE DISTRIBUTION OF THE INCIDENT IMPACT CLASSES

| Horizon (minutes) | Ground-truth Impact Class | Percentage |
|---|---|---|
| 15 | Mild | 89.6% |
| 15 | Moderate | 8.3% |
| 15 | Severe | 2.1% |
| 30 | Mild | 88.6% |
| 30 | Moderate | 9.0% |
| 30 | Severe | 2.4% |

For comparison, we also evaluate two machine learning models, random forest and XGBoost, that have been found to achieve the best performance in several comparative studies on traffic incident impact / severity prediction [1][11][17]. The machine learning models operate on the same set of features discussed earlier. They are trained using the training set consisting of 2221 incidents with ground truths. The issue of class imbalance is addressed through combined resampling (random undersampling of the majority class and the synthetic minority over-sampling technique). The parameters of the machine learning models are optimized through 5-fold grid-search cross validation.

We use two metrics, the weighted-average $F_1$ score and the macro-average $F_1$ score, to evaluate the performance of the prediction models. While the former is a commonly used metric, it is not ideal for multi-class imbalanced datasets such as the one used in this study. This is because it gives more importance to majority classes with more samples. The macro-average $F_1$ score, on the other hand, gives equal importance to all classes and is well-suited for imbalanced datasets. Therefore, we prioritize the macro-average $F_1$ score in this study and base our findings on it.

### C. Results

*Feature analysis*: In order to select an effective set of features for the experiments, we first examine the importance of the features for the impact prediction task using the random forest model. Fig. 3 shows the relative importance of all the considered features for the 15-minute prediction horizon. (The feature importances are similar for the 30-minute prediction horizon.) It can be seen that the two traffic features and the time of the incident are substantially more important than the other features. In particular, we find that the overall speed reduction ratio at prediction time, which quantifies the drop in the upstream traffic speeds in the first few minutes of the incident, is a strong indicator of the incident's future impact. Out of the incident features extracted by the LLM from the incident logs, only the number of vehicles involved and the number of lanes blocked have a reasonable degree of importance. We examine the prediction accuracies of two sets of features: One with all the features and another with only the top 5 most important features as shown in Fig. 3. We use the two machine learning models for this prediction task. Table II shows the impact prediction accuracies obtained with the two feature sets. It can be seen that in most cases, using the top 5 features yields marginally better results compared to using all the features. Therefore, we use only the top 5 features for the rest of the experiments.

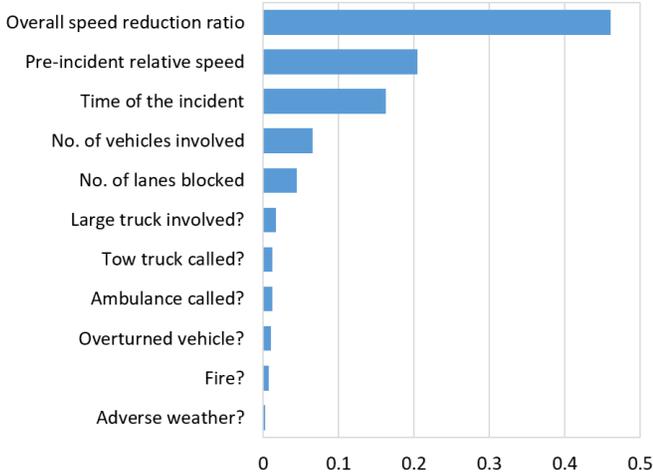

Figure 3. The relative importance of features

TABLE II. EVALUATION OF TWO FEATURE SETS USING THE MACHINE LEARNING MODELS

| Model | Horizon (min.) | Feature Set | $F_1$ Score (weighted) | $F_1$ Score (macro) |
|---|---|---|---|---|
| Random forest | 15 | Top 5 features | 0.86 | 0.59 |
| | | All features | 0.86 | 0.57 |
| | 30 | Top 5 features | 0.81 | 0.49 |
| | | All features | 0.78 | 0.44 |
| XGBoost | 15 | Top 5 features | 0.85 | 0.55 |
| | | All features | 0.85 | 0.56 |
| | 30 | Top 5 features | 0.77 | 0.49 |
| | | All features | 0.78 | 0.47 |

*Evaluation of incident impact prediction*: Table III shows the impact prediction accuracies obtained using the 3 LLMs and the 2 machine learning models on the test set with 556 traffic incidents. (All the results reported for the LLMs in this paper are the average of 3 runs.) Based on the macro-average $F_1$ score, we find that for both the prediction horizons, the best-performing LLM achieves the same or almost the same accuracy as the best-performing machine learning model that represents the current state of the art. It is worth emphasizing that the LLMs rely on just 24 examples in contrast to the machine learning models that are trained with over 2000 labelled samples. While GPT 4.1, the latest of the three LLMs evaluated, achieves the best accuracy for 15-minute-ahead prediction, Claude 3.7 Sonnet performs best for the 30-minute horizon. Among the two machine learning models, random forest achieves the best performance.

Fig. 4 shows a summary of the impact predictions made by GPT 4.1, in the form of a confusion matrix, for the 15-minute horizon in one of the experiment runs. We find that, in general, a relatively large number of mild-impact incidents are misclassified as moderate resulting in low precision and therefore a low F1 score for the moderate class, which brings down the overall macro-average F1 score. We also observe that, in general, only a small proportion of the mild-impact incidents are misclassified as severe and vice versa. Such a misclassification is more likely to be considered unacceptable by users compared to, say, a mild-impact class being misclassified as moderate.

Figure 4. The confusion matrix for 15-minute ahead prediction by GPT 4.1

TABLE III. PREDICTION RESULTS FOR TRAFFIC INCIDENT IMPACT PREDICTION

| Horizon (min.) | Model | $F_1$ Score (weighted) | $F_1$ Score (macro) |
|---|---|---|---|
| 15 | Claude 3.7 Sonnet | 0.83 | 0.53 |
| | Gemini 2.0 Flash | 0.87 | 0.56 |
| | GPT 4.1 | 0.87 | 0.59 |
| | Random Forest | 0.86 | 0.59 |
| | XGBoost | 0.85 | 0.55 |
| 30 | Claude 3.7 Sonnet | 0.79 | 0.48 |
| | Gemini 2.0 Flash | 0.78 | 0.47 |
| | GPT 4.1 | 0.78 | 0.45 |
| | Random Forest | 0.81 | 0.49 |
| | XGBoost | 0.77 | 0.49 |

We investigate the effect of the number of examples in the system prompt on the performance of the LLMs by varying the value of $k$ in the example selection method. As we have empirically set the value of $m$ as 12, setting the value of $k$ as 0,1,2 and 3 will result in the final system prompt having 0, 12, 24 and 36 examples, respectively. The macro-average F1 score for each of these numbers of examples are shown in Table IV. In all the cases, increasing the number of examples from 0 to 12 yields a significant improvement in the macro-average F1 score. Increasing the number of examples from 12 to 24 improves the accuracy significantly in most instances of 15-minute-ahead prediction. However, an increase from 24 to 36 examples do not yield a significant improvement in most cases. It is based on this analysis that we have decided to use 24 examples in the final system prompt (i.e., $k$ is set to 2).

TABLE IV. EVALUATION OF DIFFERENT NUMBER OF EXAMPLES IN THE SYSTEM PROMPT

| Horizon (min.) | Model | $F_1$-macro 0 examples | $F_1$-macro 12 examples | $F_1$-macro 24 examples | $F_1$-macro 36 examples |
|---|---|---|---|---|---|
| 15 | Claude 3.7 Sonnet | 0.50 | 0.53 | 0.53 | 0.54 |
| | Gemini 2.0 Flash | 0.36 | 0.53 | 0.56 | 0.56 |
| | GPT 4.1 | 0.46 | 0.51 | 0.59 | 0.59 |
| 30 | Claude 3.7 Sonnet | 0.45 | 0.48 | 0.48 | 0.47 |
| | Gemini 2.0 Flash | 0.37 | 0.47 | 0.47 | 0.49 |
| | GPT 4.1 | 0.41 | 0.45 | 0.45 | 0.47 |

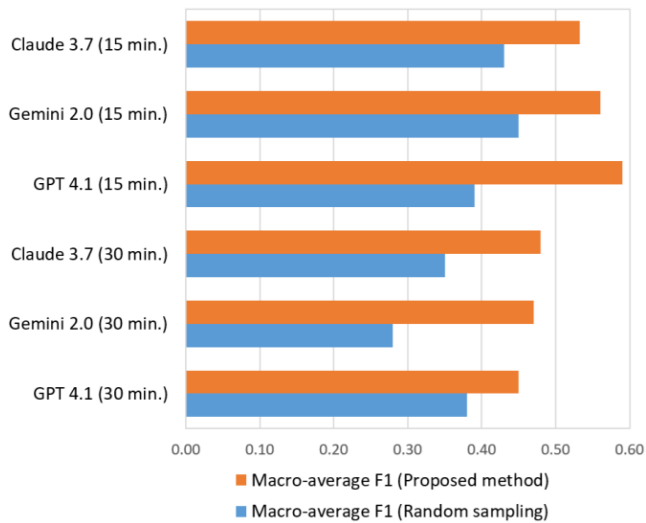

Figure 5.  Comparison of the proposed example selection method and random sampling

In order to validate the proposed method of selecting examples to be included in the system prompt, we compare it with the baseline of randomly selecting the same number (24) of examples from the training set. Figure 5 shows this comparison where the proposed example selection method results in a substantial and consistent improvement over using randomly-selected examples in the system prompt.

## VI. CONCLUSIONS

In this study, we have investigated the novel idea of using LLMs to directly forecast the impact of traffic incidents. For this, we have proposed an LLM-based solution which uses a method of selecting effective examples to instruct the LLM about the prediction task. We have evaluated the performance of 3 advanced LLMs for incident impact prediction on a large dataset of real traffic incidents and compared them with two machine learning models that are known to be the most effective ones. We find that for each prediction horizon, the prediction accuracies of the best-performing LLM and the best-performing machine learning model are comparable, despite the former not being trained or fine-tuned for incident impact prediction. Among the evaluated LLMs, GPT 4.1 and Claude 3.7 Sonnet attain the best performance for 15-minute and 30-minute prediction horizons, respectively, with Gemini 2.0 Flash being the second-best model. The results also show that the proposed example selection method enables the LLMs to produce significantly better results compared to the common practice of using a random sample of examples. Overall, this study attests to the feasibility of using LLMs to forecast the impact of traffic incidents.

There are a few ways of further extending this work. Our LLM-based solution relies only on a small number of features. The judicious addition of more features, including contextual ones, could possibly improve the performance further. Also, more sophisticated prompt optimization strategies (e.g., [18]) could potentially yield better results. The usefulness of providing the LLM with domain-specific expert knowledge on traffic incident scenarios could be investigated. With the capabilities of LLMs advancing at a very rapid pace, it is reasonable to suppose that in the near future, LLMs would be able to clearly outperform conventional methods of predicting the impact of traffic incidents.